\documentclass[lettersize,journal]{IEEEtran}
\usepackage{amsmath,amsfonts}
\usepackage{stfloats}
\usepackage{url}
\usepackage{graphicx}
\usepackage{cite}
\usepackage{booktabs}
\usepackage{multirow}
\newcommand\para[1]{{\vspace{5pt} \bf \noindent #1 \hspace{3pt}}}
\newenvironment{packed_itemize}{
	\begin{list}{\labelitemi}{\leftmargin=1.em}
		\setlength{\itemsep}{1pt}
		\setlength{\parskip}{0pt}
		\setlength{\parsep}{0pt}
		\setlength{\headsep}{0pt}
		\setlength{\topskip}{0pt}
		\setlength{\topsep}{0pt}
		\setlength{\partopsep}{0pt}
}{\end{list}}

\begin{document}

\title{Video-Based Palm-Vein Authentication under Challenging Conditions}

\author{Xiaofeng~Yan, Kechen~Liu, Abhilash~Venkatesh, Cathy~Zhang, Xia~Zhou, and Salvatore~Stolfo%
\thanks{This work was performed at the Department of Computer Science, Columbia University, New York, NY 10027 USA.}%
\thanks{X. Yan, X. Zhou, and S. Stolfo are with the Department of Computer Science, Columbia University, New York, NY 10027 USA.}%
\thanks{K. Liu is with Princeton University, Princeton, NJ 08544 USA.}%
\thanks{C. Zhang is with Stanford University, Stanford, CA 94305 USA.}%
\thanks{A. Venkatesh contributed to this work while at Columbia University and is currently with Amazon.com Services LLC, USA.}%
\thanks{Corresponding author: Salvatore Stolfo (e-mail: sal@cs.columbia.edu).}}

\markboth{Preprint}%
{Yan \MakeLowercase{\textit{et al.}}: Video-Based Palm-Vein Authentication under Challenging Conditions}

\maketitle
\raggedbottom

\begin{abstract}
Palm-vein biometrics are increasingly used for secure, contactless authentication. Yet real-world deployment exposes them to surface noise (sweat, dirt), illumination and motion variation, and temperature-driven changes in vascular visibility, which remain underexplored for lack of data captured under such conditions.
To study these effects, we introduce the Columbia University Palm-vein (CUP) dataset, to our knowledge the first public video-based palm-vein dataset. CUP records every palm under four surface conditions (a clean baseline, warm, wet, and dirty) and pairs each subject with physiological and demographic metadata.
On it we benchmark twenty-one recognizers spanning static, video, and multi-frame aggregation architectures. Models that verify reliably on clean palms lose most of their accuracy on dirty ones, and the mean equal error rate (EER) roughly quadruples.
We recover much of that robustness along both axes of the capture.
Temporally, a consensus over the few frames the sensor already returns cancels transient corruption; spatially, a test-time matcher that adds no learned parameters fuses the global cosine with a saliency-steered region-level optimal transport that routes the comparison around corrupted regions.
The full design leads on every surface of CUP in EER, TAR@FAR${=}0.01$, and Rank-1, at $4.3$M parameters and $3.1$ GFLOPs, a fraction of the video models' cost.
Attached to four frozen state-of-the-art backbones it cuts their mean EER by $29$--$37\%$ without retraining, and on four public single-image datasets the regional matching alone still helps.
A preliminary audit across ten demographic and physiological traits finds two warm-condition gaps, along body water and gender, that survive multiple-comparison correction.
CUP will be released for non-commercial research use at \url{https://github.com/MobileX-CU/CUP_v1} upon publication.
\end{abstract}

\begin{IEEEkeywords}
Palm-vein recognition, biometrics, video-based authentication, optimal transport, robustness, dataset, fairness.
\end{IEEEkeywords}

\section{Introduction}
\label{sec:intro}
\IEEEPARstart{P}{alm}-vein biometrics are attracting growing interest, offering unique advantages over conventional modalities such as fingerprints or the iris. Vein patterns are internal and visible only under near-infrared (NIR) illumination, making them difficult to observe or capture covertly~\cite{watanabe2008palm,kumar2008hand,zhou2011human,wu2020reviewpalmvein}. The contactless capture is also hygienic, boosting user acceptance in the post-pandemic era. Beyond security and hygiene, the technology offers competitive accuracy with inexpensive hardware~\cite{jain2004introduction}. The vascular pattern also carries high biological entropy: it stays unique even for identical twins, where face recognition struggles. These advantages have spurred rapid commercial adoption by industry leaders such as Amazon and Alipay.

\begin{figure}[t]
\centering
\includegraphics[width=\columnwidth]{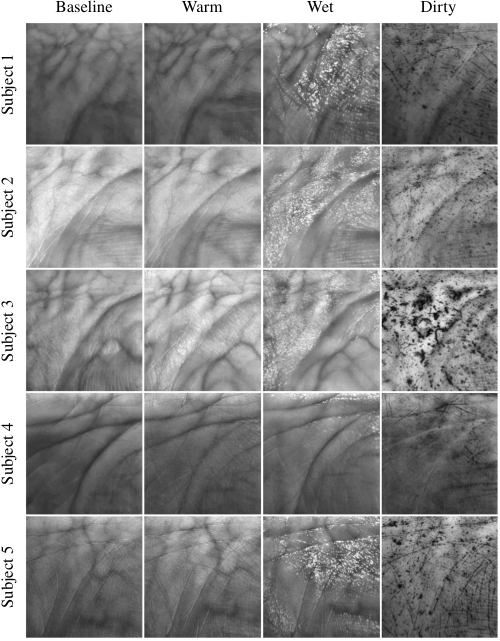}
\caption{\textbf{Surface Degradation in CUP.} Columns are a clean baseline and the three degrading surface conditions (warm, wet, and dirty) that corrupt the near-infrared vein signal; rows are different subjects. Every palm is re-recorded under all four conditions.}
\label{fig:surfaces}
\end{figure}

Despite this success, the technology's performance under non-ideal, real-world conditions is not well understood. The non-planar topography of the palm makes static features sensitive to acquisition artifacts: pose variation, shadows, and surface contaminants like sweat or dirt (Fig.~\ref{fig:surfaces}). These artifacts degrade matching performance. While industry solutions claim high accuracy, their proprietary data hinders independent validation. On the other hand, public datasets have been developing fast, from contact plates~\cite{kabacinski2011put} and controlled contactless capture~\cite{tome2015vulnerability} to multi-wavelength imaging~\cite{kauba2019combined}, two-session collection~\cite{zhang2018palmprint}, and large-scale, weakly cooperative capture~\cite{luo2024palm}. However, even the least controlled of them release only still images of clean palms, leaving the surface conditions of everyday use untested.

We built CUP to supply exactly these conditions; to our knowledge it is the first public video-based palm-vein dataset. The format is video for two reasons. A contactless capture natively returns a short sequence as the user positions the hand over the sensor, so the frames come for free. They also carry complementary identity evidence, which has proved valuable for face recognition under noisy, less controlled capture~\cite{yang2017nan,kim2022caface}. The dataset provides $5{,}049$ curated two-second NIR clips from $109$ subjects, recorded under a protocol that mirrors everyday use and released for non-commercial research use. CUP loosens control along two further axes. Each palm is presented with the natural pose and distance variation of routine interaction and is re-recorded under four surface conditions: a clean baseline, warm, wet, and dirty. Every subject is paired with demographic and physiological metadata (\S\ref{sec:dataset}).
We then use CUP to quantify how everyday degradation affects recognition. The drop is severe and universal: recognizers that verify reliably on clean palms lose most of their accuracy on dirty ones. The twenty-one recognizers we benchmark span vascular-specific pipelines, generic image backbones, end-to-end video models, and multi-frame aggregation. Across all of them, the mean equal error rate (EER) roughly quadruples, from $4.11\%$ to $17.35\%$. Training on degraded data and averaging over frames help only partially: on dirty palms every recognizer stays far above its own clean-palm floor (\S\ref{subsec:lab_to_wild_gap}).

Given degradation of this severity, the global embedding is itself corrupted. Water reflections or surface contaminants can wipe out whole regions of the vein signal at once, and the damage pools into the single vector being compared. We instead compare palms with a matcher that operates along both dimensions of a video capture, space and time. Person re-identification and face recognition survive occlusion the same way: the reliable parts of the input decide the match~\cite{miao2019pgfa,gao2020pvpm,nguyen2025emd,song2019occlusion}. Spatially, we compare palms region by region, so that the regions that survive the degradation can decide the match. The comparison runs as an entropic optimal transport, which shifts matching weight off corrupted regions. A saliency prior then spends that weight on the regions that best identify each gallery. We keep the global embedding alongside, because some probes fail the other way: their local patches are misleading while the overall pattern still matches. Fusing the two views lets each recover the probes the other loses. Temporally, a consensus over the frames of a clip cancels transient corruptions and recovers the stable vein signal beneath. The only training we add is a light consensus objective that makes each frame reliable on its own, leaving the architecture untouched.

Extensive experiments support this design. On CUP it leads on every surface condition and every metric, with the widest margin on the hardest, dirty condition. Because the matcher adds no trainable parameters, the gain is not tied to our encoder. Attached to four frozen CNN and transformer backbones, it improves every one on every surface and metric without retraining; it also generalizes to four public single-image datasets~\cite{hao2008multispectral,luo2024palm,zhang2018palmprint,tome2015vulnerability}. Under a subject-disjoint protocol with no shared subjects, the ordering holds and the margin widens (\S\ref{subsec:main-exp}). We further probe where the value of video lies. A few frames sampled evenly across the two-second capture already realize the full multi-frame gain, whereas consecutive frames are largely redundant. The benefit thus comes from covering the capture window rather than from a high frame rate, and it adds no acquisition cost.

Finally, we use CUP's metadata to study systematic performance differences across user groups. During collection we observed that vein visibility varies across participants, so CUP records per-subject demographic (e.g., age, gender) and physiological (e.g., BMI, body fat) measurements. We audit recognition accuracy across ten such traits. The audit finds two warm-condition differences, along body water and gender, that survive multiple-comparison correction; they mark where a larger study should look.

In summary, this paper makes three contributions:
\begin{itemize}
\item \textbf{Dataset and benchmark.} We release CUP, to our knowledge the first public video-based palm-vein dataset with four surface conditions and paired metadata. With it we benchmark twenty-one recognizers under a matched protocol, quantifying this drop and tracing it to pooled embeddings.
\item \textbf{A matcher built on what the benchmark reveals.} Guided by the failure mode CUP exposes, we propose a dual-view encoder with a region-level optimal-transport matcher that adds no learned parameters. The full method restores accuracy on CUP, and the matcher alone carries its gain to other backbones and public datasets without retraining.
\item \textbf{A dataset-enabled fairness signal.} Enabled by CUP's per-subject metadata, we conduct a preliminary audit of recognition accuracy across ten demographic and physiological traits, which we report as a preliminary signal rather than a characterization of palm-vein fairness.
\end{itemize}

\section{Related Work}
\label{sec:related}

\subsection{Contactless Palm-Vein Datasets}
\label{ssec:rw-data}

Public palm-vein datasets have grown steadily in scale and in the conditions they capture. Palm veins were first imaged as the near-infrared band of multispectral palmprint collections such as CASIA-MS~\cite{hao2008multispectral} and PolyU-MS~\cite{zhang2010multispectral}, which image each palm under several illumination wavelengths in a constrained setup. Dedicated palm-vein databases then appeared under controlled contactless capture: VERA~\cite{tome2015vulnerability} pairs $2{,}200$ images of $220$ palms with a spoofing protocol for vulnerability analysis. Later collections each widened one axis of variation. TJ-PV~\cite{zhang2018palmprint} records twenty images per palm for $600$ palms across two sessions, adding the time gap that enrolled systems face; PLUSVein~\cite{kauba2019combined} images palmar and finger veins at multiple near-infrared wavelengths; and FYO~\cite{toygar2020fyo} pairs palmar, dorsal, and wrist views of the same subjects. The largest to date is SCUT~\cite{luo2024palm}, with $11{,}000$ images of $1{,}100$ palms from $550$ subjects under unconstrained, weakly cooperative capture. It has become the standard large-scale training source for the modality.

Across all of these, however, surface degradation is not included, though it has long been studied in fingerprint benchmarks~\cite{maio2004fvc} and recently in in-the-wild palmprint~\cite{seyedmohammadi2026xpalm}. The closest same-modality effort remains small, at $500$ images from $100$ subjects~\cite{chate2025condition}. CUP therefore adds a new dimension for palm veins: to our knowledge it is the first public palm-vein dataset to release the near-infrared video clips themselves, recorded under real surface degradation and paired with demographic and physiological metadata.

\subsection{Palm-Vein Recognition}
\label{ssec:rw-recognition}

Most palm-vein recognizers represent a capture with a single holistic descriptor. Early methods extracted handcrafted features from static NIR images using local descriptors such as Gabor filters, Local Binary Patterns, SIFT, and the Radon transform~\cite{mirmohamadsadeghi2014palm,han2012palm,ladoux2009palm,zhou2011human}. The shift to deep learning improved discriminability, and Jia et al.~\cite{jia2020cnneval} systematically benchmarked CNNs for palmprint and palm-vein recognition. Later work addressed data scarcity through few-shot learning~\cite{marattukalam2021nshot} and contrastive learning~\cite{ma2023focal}, and pursued stronger discrimination through attention mechanisms~\cite{htet2023contactless} and margin- or quality-aware objectives. MDNet~\cite{kuzu2021loss} adds the ArcFace margin~\cite{deng2019arcface}; AMPVNet~\cite{luo2024palm} adds domain-specific augmentation and an adaptive-margin loss. Across these advances, matching ultimately reduces the feature map to a single pooled embedding that is compared by cosine similarity.

A complementary line of work divides the ROI image or the feature map into regions, injecting local vein structure into training. Nayar et al.~\cite{nayar2021graph} partition the palm into variable-size blocks that form graph nodes, linked wherever a vein pattern spans them; block descriptors have been fused with deep features for verification~\cite{elghandour2021block}; and multiscale transformers couple patch tokens by self-attention to mix local and global cues~\cite{qin2023transformer}. More recently, RSNet~\cite{luo2025rsnet} argues that generic patch division ignores palm-vein structure and instead partitions the feature map using physiological priors on palm shape and vein layout.

Both lines of work reduce each capture to one pooled embedding before matching. When part of the surface is degraded, the pooling itself becomes the bottleneck. Averaging every spatial location into one vector lets a few corrupted regions (a wet smear, a patch of dirt) drag down the similarity to the genuine gallery even when most of the pattern is intact. Degraded palm-vein data is also still too scarce for training alone to close the gap. Region-aware training helps only indirectly, since the regional structure is collapsed again at comparison time. Letting the reliable regions decide the match at matching time remains much less explored.

\subsection{Matching under Partial Corruption}
\label{ssec:rw-challenging}

One family of methods attacks partial corruption spatially, representing a sample as a set of local descriptors so that the reliable regions can decide the match. Optimal transport (OT) gives this idea a principled form. The Earth Mover's Distance was introduced to compare such sets for image retrieval~\cite{rubner2000emd}; entropic regularization then made it fast and differentiable through the Sinkhorn algorithm~\cite{cuturi2013sinkhorn,peyre2019computational}. The machinery has since spread across recognition. Person re-identification matches only the parts visible under occlusion, aligning them by pose or comparing them by EMD~\cite{miao2019pgfa,gao2020pvpm,he2018partial,wang2020highorder,nguyen2025emd}, or weighting them by a learned quality~\cite{wang2023qpm}. Face recognition learns masks that discard corrupted feature elements~\cite{song2019occlusion,qiu2022from} and re-ranks retrieved candidates by patch-wise EMD~\cite{phan2022deepfaceemd}. Few-shot recognition matches dense regions through DeepEMD~\cite{zhang2020deepemd}, with SuperGlue~\cite{sarlin2020superglue} doing the same for sparse keypoints.

Yet for palm veins this idea remains much less explored, and the cues it relies on elsewhere are missing. A contactless palm-vein ROI has no canonical part layout and no external oracle for which vein regions are trustworthy. Its degradation is a diffuse attenuation of the vein texture rather than a discrete occluder to detect, so the transport itself must discover which regions to trust.

A second family attacks the problem temporally, exploiting the fact that a contactless capture is itself a short video. It splits into two lines. Multi-frame aggregation, most developed in video-based face recognition, scores frames by quality and pools them into one template: neural aggregation~\cite{yang2017nan}, component-wise aggregation~\cite{gong2019cfan}, set pooling with ghost clusters~\cite{zhong2018ghostvlad}, feature-aggregation networks~\cite{liu2019fan}, learned frame weighting~\cite{rao2017attention}, and blur-robust ensemble features~\cite{ding2018trunk}. End-to-end video models instead read motion directly, from convolutional spatio-temporal architectures~\cite{tran2015c3d,tran2018r2plus1d,hara2018can,wang2016tsn,carreira2017quo} to video transformers~\cite{liu2022videoswin,li2022mvitv2,bertasius2021space,arnab2021vivit}, at several times the parameters of a single-image recognizer. For palm veins, both lines meet the same obstacles. The aggregation methods pool the frames into one descriptor before matching, reintroducing the pooling that falters under partial degradation; their frame-quality scoring is also learned for face appearance. Motion offers little either: unlike facial motion, hand movement under near-infrared is subtle and blurs an already low-contrast pattern. The motion models thus spend their capacity on dynamics a palm-vein burst does not contain, leaving their value at denoising.

We instead keep a strong global recognizer and add a region-level transport at matching time, steered by the intrinsic saliency of the palm's own vein structure. The frames of the capture are combined at this stage rather than pooled beforehand, so the reliable regions decide the match when part of the input is corrupted (\S\ref{sec:method}).

\section{Video-Based Palm-Vein Imaging Dataset}
\label{sec:dataset}
\begin{figure}[t!]
\centering
\includegraphics[width=\columnwidth]{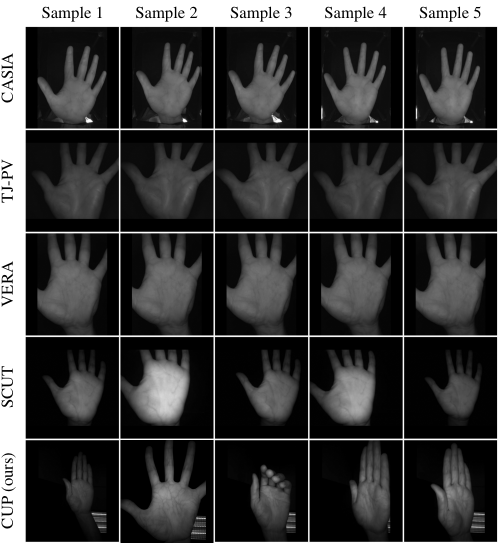}
\caption{\textbf{CUP versus Public Palm-Vein Datasets.} Each row is one dataset: four public sets (CASIA~\cite{hao2008multispectral}, TJ-PV~\cite{zhang2018palmprint}, VERA~\cite{tome2015vulnerability}, SCUT~\cite{luo2024palm}) and CUP, five near-infrared samples of a single subject each; the intra-subject variability is visibly larger for CUP. See Table~\ref{tab:datasets} for detailed statistics of each dataset.}
\label{fig:datasets}
\end{figure}

We introduce CUP to understand how recognition behaves once palms are presented in challenging, everyday settings. To the best of our knowledge, it is the first public video-based palm-vein dataset.
It captures several of the stressors that real deployments impose, with intra-subject variability far beyond prior sets. It complements prior, largely single-image benchmarks with continuous video, controlled surface degradation, and paired physiological and demographic metadata.
Cross-dataset sample comparisons are shown in Fig.~\ref{fig:datasets}.
Each palm is recorded as a continuous near-infrared video stream, from which we cut two-second clips and keep those that pass curation (\S\ref{subsec: roi}). Capturing a two-second clip rather than a single frame adds no hardware cost, since the extra frames are simply what the sensor already records as the palm is presented. The protocol is designed to reflect everyday use. It accommodates the natural pose variation (e.g., finger extension and palm bending) and distance changes (5--30\,cm) that arise in routine interaction. It also applies four surface conditions as controlled surrogates of everyday stressors: a clean baseline, warm (temperature-induced vasodilation that alters vein visibility), wet (water reflections), and dirty (surface occlusions from contaminants). Representative frames are shown in Fig.~\ref{fig:surfaces}.
In total, CUP provides 5{,}049 curated two-second clips from 109 subjects. Each subject is accompanied by physiological and demographic metadata (including age, gender, body mass index (BMI), body fat, SpO\textsubscript{2}, and pulse rate). The dataset will be released for non-commercial research use, under a data-use agreement, upon publication.
\subsection{Data Acquisition}

\para{Hardware Setup}
Palm-vein videos were recorded using a mobile camera operating at 30 frames per second with a resolution of 1920$\times$1080 pixels. To enhance vascular visibility, a cut-off filter was installed to block visible wavelengths below 700\,nm. Three NIR LEDs operating at 850\,nm illuminated the palm.

\para{Data Collection Process}
Our data collection protocol was approved by the Institutional Review Board of Columbia University. We recruited 109 subjects, each of whom provided informed consent prior to participation. All subjects were compensated \$10 for their time. For each subject, we first collected the following demographic and physiological information: age, gender, and self-reported race; height, weight, BMI, body fat percentage, muscle mass, and basal metabolic rate, all measured with a commercial smart scale~\cite{ge_fit_plus}; and SpO\textsubscript{2}, pulse rate, respiration rate, pleth variability index, and perfusion index, measured with an FDA-approved pulse oximeter~\cite{masimo_mightysat}.
Next, each subject placed a palm above our hardware for video capture. Every capture follows the same movement protocol: the subject rotates the palm from 0° (facing the camera) to approximately 60° and moves it vertically between 5 and 30 cm, covering the placements and standoffs of routine interaction. We repeat that protocol under four surface conditions.
\begin{packed_itemize}
    \item \textbf{Baseline}: The palm is clean, dry, and at normal temperature.
    \item \textbf{Warm Palm}: The palm is warmed to temperatures exceeding 38.5$^\circ$C using a heat pad to simulate elevated peripheral circulation.
    \item \textbf{Wet Palm}: The palm is sprayed with water to simulate moisture from sweat or environmental humidity.
    \item \textbf{Dirty Palm}: Soil is gently applied to the palm surface to introduce realistic occlusions and contaminants.
\end{packed_itemize}
Because the movement protocol runs inside every condition, each clip already carries pose and standoff variation. The surface effects we report are therefore measured on top of that geometric variation rather than in a fixed pose.

\subsection{Data Processing}
\label{subsec: roi}
Unconstrained video capture introduces challenges absent from static, controlled datasets. Traditional heuristic ROI methods that rely on inter-finger valley points~\cite{obayya2020contactless} fail when the fingers
are not well separated or the viewpoint is unfavorable. We instead detect 21 hand landmarks per frame using a deep-learning-based hand tracker~\cite{mediapipe_hands}.
From the 21 landmarks, we select four anatomically stable anchor points: $A$ and $B$ at the metacarpophalangeal joints of the index and little fingers, $C$ at the carpometacarpal joint of the thumb, and $D$ at the wrist joint.
These landmarks are chosen because they correspond to rigid skeletal joints, which offer substantially greater stability against soft-tissue deformation than skin-surface points. Although the quadrilateral formed by $A$-$B$-$D$-$C$ covers the palm, the regions near the thumb base ($C$) and the wrist ($D$) remain subject to non-rigid deformation during natural hand movements. To obtain a stable, planar region suitable for perspective warping, we construct an inset quadrilateral with vertices $A$, $B$, $F$, and $E$.
Points $E$ and $F$ are obtained by linear interpolation along the side segments of the palm so as to exclude the unstable regions:
\begin{equation}
    E = C + 0.25 \cdot (A - C), \quad F = D + 0.25 \cdot (B - D).
\end{equation}
The final ROI is defined by the polygon $A$-$B$-$F$-$E$. We apply a perspective transformation to warp this region into a canonical square image for downstream processing.
To mitigate the high-frequency jitter of independent single-frame detections, we apply a moving-average filter to the raw landmark coordinates $\mathbf{p}_t$ over a window of $2k{+}1$ frames ($k{=}2$ in all experiments), yielding the smoothed coordinates $\hat{\mathbf{p}}_t$:
\begin{equation}
    \hat{\mathbf{p}}_t = \frac{1}{2k+1} \sum_{i=-k}^{k} \mathbf{p}_{t+i}.
\end{equation}
This operation acts as a low-pass filter on the ROI boundaries, removing sensor quantization noise and small detection errors. It stabilizes spatial alignment across frames, so downstream matching operates on consistent ROIs rather than on tracking jitter.
After ROI extraction, the clips pass through a two-stage curation pipeline: an automated screening stage that computes per-frame
quality metrics (skewness, aspect ratio, shadow, and entropy), followed by manual review.
\subsection{Data Analysis}
\begin{figure}[t!]
    \centering
    \includegraphics[width=\linewidth]{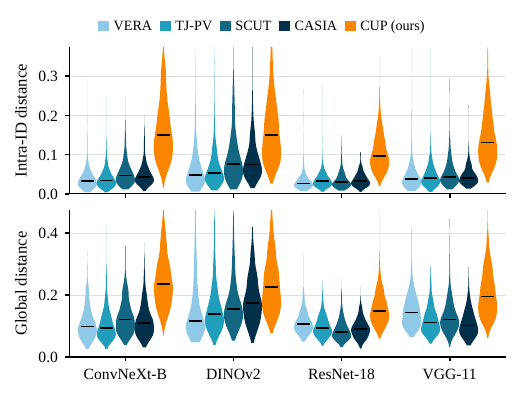}
    \caption{\textbf{Feature Dispersion of CUP versus Public Datasets.} Per-sample intra-identity distance (top) and global distance (bottom), grouped by off-the-shelf feature extractor. Under every backbone, CUP (orange) is far more dispersed than the public sets (blue gradient).}
    \label{fig:feat}
\end{figure}
\begin{figure*}[t!]
\centering
\includegraphics[width=\textwidth]{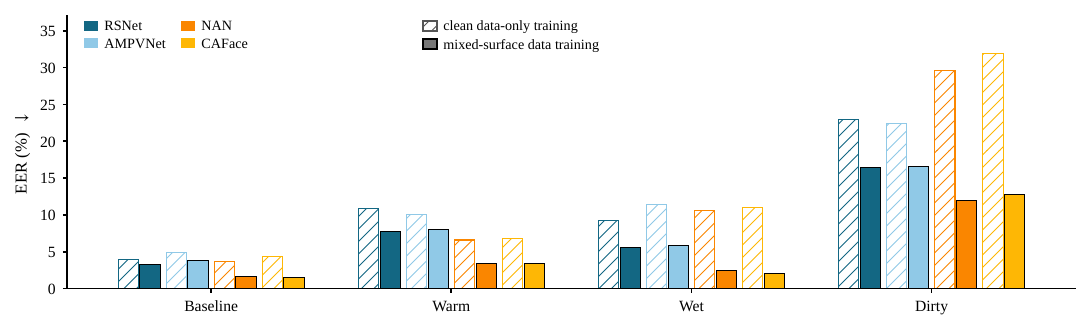}
\caption{\textbf{The Lab-to-Wild Gap on CUP.} Open-set EER per surface for two static (RSNet~\cite{luo2025rsnet}, AMPVNet~\cite{luo2024palm}; blue) and two multi-frame (NAN~\cite{yang2017nan}, CAFace~\cite{kim2022caface}; orange) recognizers, trained on clean-only (hatched) or mixed-surface data (solid). Every recognizer degrades sharply toward dirty; mixed-surface training lowers error but leaves a large residual above the baseline floor.}
\label{fig:degradation}
\end{figure*}
\label{subsec:stats}

\para{Intra-Subject Diversity} We first measure how large CUP's intra-subject variability is relative to existing benchmarks. For each dataset, we embed every sample with four pre-trained backbones (ConvNeXt-B~\cite{liu2022convnet}, DINOv2~\cite{oquab2023dinov2}, ResNet-18~\cite{he2016deep}, VGG-11~\cite{simonyan2014very}) and characterize its feature geometry with two per-sample distances (Fig.~\ref{fig:feat}). For a sample with $\ell_2$-normalized feature $\mathbf{x}$, let $\boldsymbol{\mu}_{\mathrm{id}}$ denote the centroid of its own identity (the mean feature over all captures of that palm) and $\boldsymbol{\mu}$ the global centroid (the mean over all samples in the dataset). The intra-ID distance $\lVert\mathbf{x}-\boldsymbol{\mu}_{\mathrm{id}}\rVert$ measures how far repeated captures of the same palm scatter around their own prototype (within-identity variability). The global distance $\lVert\mathbf{x}-\boldsymbol{\mu}\rVert$ instead measures how widely the dataset spreads over the feature space as a whole. On the intra-ID measure, the within-identity dispersion of CUP is $2$--$3\times$ that of every public set under every backbone (Fig.~\ref{fig:feat}, top). Repeated captures of one palm scatter far more widely in feature space, which is the direct signature of real-world surface, pose, and distance variation. The global measure guards against a trivial reading: the extra scatter is not attributable to a single loose identity but persists relative to the dataset-wide mean (Fig.~\ref{fig:feat}, bottom). Together, the two measures indicate reduced cluster compactness: the median intra-ID distance of CUP is ${\approx}0.65$ of its median global distance, versus ${\approx}0.38$ for the public sets. Identities thus occupy a far larger share of the populated feature space and are correspondingly harder to separate.

\para{Curation and Statistics} Curation operates at the palm level: a palm whose clips are all rejected leaves the dataset, so the 109 subjects yield 210 palms rather than 218. After the two-stage curation pipeline, the open-set protocol splits the dataset into 2{,}376 training clips and 2{,}673 test clips with no palm-level overlap between the splits. Per-condition counts are approximately balanced: 1{,}288 Baseline, 1{,}234 Wet, 1{,}276 Dirty, and 1{,}251 Warm clips. The per-palm clip count has a median of 25, reflecting variable retention from quality filtering. Retention also varies by hand, reflecting a handedness effect. Most participants are right-handed and produced steadier right-hand videos during the pose-variation protocol, yielding higher post-filter retention for right palms (median 27 versus 22.5 clips per palm).
\subsection{The Lab-to-Wild Gap: Standard Training Is Not Enough}
\label{subsec:lab_to_wild_gap}
CUP allows us to measure directly how well standard recognition performs once palms are presented in everyday, degraded conditions. We benchmark representative static and multi-frame recognizers, each trained in two ways: on clean (baseline) data alone and on the full mixed-surface set that spans all four conditions. We then evaluate them per surface under the open-set protocol detailed in \S\ref{sec:experiments}. We select four representative models: the two best static recognizers (RSNet~\cite{luo2025rsnet}, AMPVNet~\cite{luo2024palm}) and the two best multi-frame recognizers (NAN~\cite{yang2017nan}, CAFace~\cite{kim2022caface}), shown in Fig.~\ref{fig:degradation}. The full benchmark is reported in \S\ref{sec:experiments}.

\para{Degradation is severe and universal.} Every recognizer degrades sharply from clean to dirty. A strong static model such as RSNet rises from an EER of $3.3\%$ on clean palms to $16.5\%$ on dirty ones. The lowest dirty-condition EER any recognizer in \S\ref{sec:experiments} reaches is still about $12\%$. Standard training offers two mitigations: mixing degraded surfaces into the training data and aggregating the frames of a clip. Both help only partially (Fig.~\ref{fig:degradation}). Mixed-surface training lowers error on every degraded surface relative to clean-only training. The two multi-frame recognizers, which aggregate the frames of each clip, benefit the most. Trained on mixed surfaces, they stay below the static models on every surface and hold their EER below $3.4\%$ on the baseline, warm, and wet conditions. Frame aggregation alone is not enough: trained on clean data only, they degrade even more sharply on dirty palms than the static models do. Neither measure closes the gap: on dirty palms every recognizer remains far above its own baseline-condition floor. Standard training therefore absorbs the mild degradations but not the severe ones.

\section{Method}
\label{sec:method}

\begin{figure*}[t!]\centering
  \includegraphics[width=0.80\textwidth]{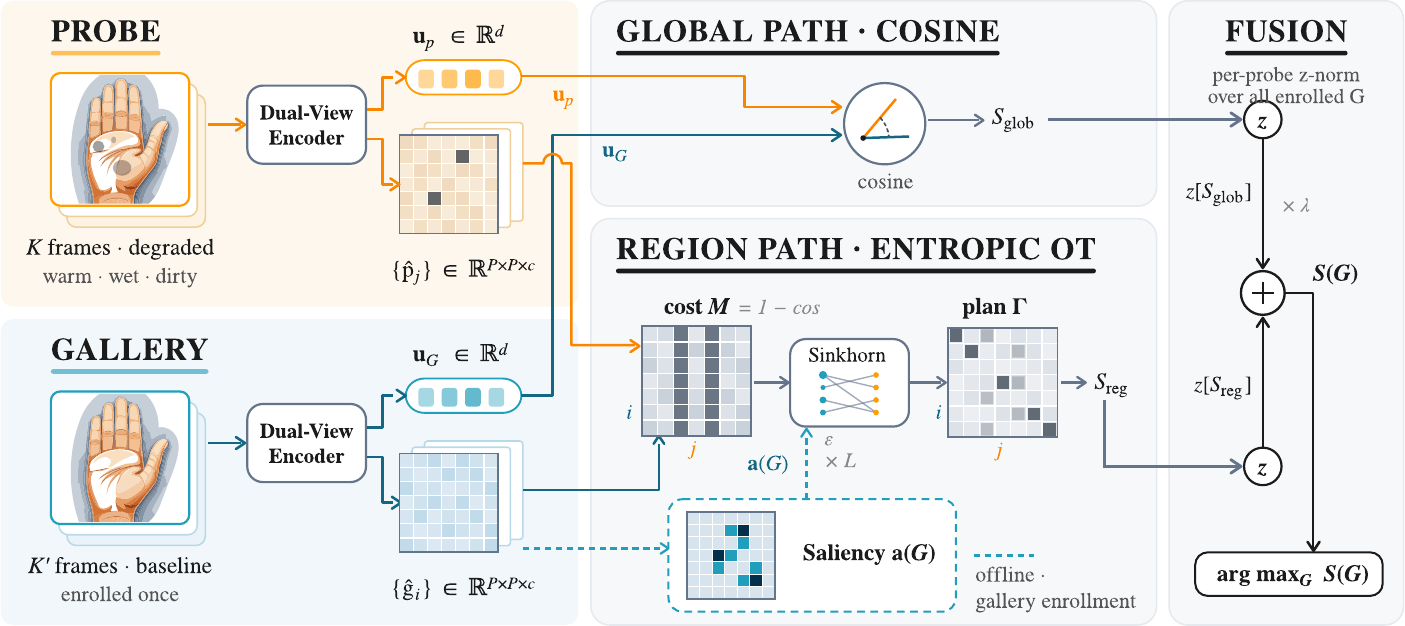}
  \caption{\textbf{Method Overview.} A shared Dual-View Encoder (left) turns the probe clip and each enrolled
  gallery clip into a global embedding and a grid of regional features. From these, the test-time
  matcher scores a probe against each gallery: a region path (lower center) matched by entropic optimal
  transport, a global path (upper center) compared by cosine similarity, and a complementary fusion (right) that
  yields the final identity decision.}
  \label{fig:pipeline}
\end{figure*}

\subsection{Dual-View Encoder}
\label{subsec:multiframe}
The encoder turns each input clip into two descriptors: a global embedding for holistic comparison, and a
$P\times P$ grid of region descriptors that preserves where on the palm each feature comes from
(Fig.~\ref{fig:pipeline}). A query is a probe clip of $K$ near-infrared frames, matched in the open-set regime
against a gallery of identities, each enrolled once from a short clip of $K'$ frames ($K'{=}1$ for a single
template). Both sides pass through the same encoder, so the surviving regions remain available to the
matcher when the pooled summary is compromised. A frozen backbone $f$ maps each frame, in a single forward pass,
to a $P\!\times\!P$ feature map. The region view keeps that map as $R{=}P^{2}$ region descriptors
$\{\mathbf{r}^{(t)}_j\}\!\in\!\mathbb{R}^{c}$; the global view collapses it by global average pooling (GAP) into
a pooled feature $\mathbf{f}^{(t)}\!\in\!\mathbb{R}^{c}$. We reduce a clip to one descriptor per view by
averaging over the $K$ frames and $\ell_2$-normalizing, region-wise for the region view and through the
projection head $\phi:\mathbb{R}^{c}\!\to\!\mathbb{R}^{d}$ for the global view:
\begin{equation}
  \hat{\mathbf{p}}_{j}=\overline{\tfrac1K\textstyle\sum_{t}\mathbf{r}^{(t)}_{j}},
  \qquad
  \mathbf{u}_{p}=\overline{\phi(\bar{\mathbf{f}})},
  \qquad
  \bar{\mathbf{f}}=\tfrac1K\textstyle\sum_{t}\mathbf{f}^{(t)},
  \label{eq:tsn}
\end{equation}
where $\overline{(\cdot)}$ denotes $\ell_2$ normalization and, since $\phi$ is linear, $\mathbf{u}_{p}$ equally
reads as the renormalized average of the per-frame embeddings $\phi(\mathbf{f}^{(t)})$ (sparse temporal sampling in the spirit of TSN~\cite{wang2016tsn}). Averaging cancels the transient, frame-local corruption while
the stable vein structure accumulates. It needs no training and is applied identically on both sides, giving the
aggregated descriptors $(\mathbf{u}_p,\{\hat{\mathbf{p}}_j\})$ and $(\mathbf{u}_G,\{\hat{\mathbf{g}}_i\})$ that the
matcher consumes. The frame count and the sampling scheme are studied in \S\ref{subsec:multiframe-exp}.

We want the average in~\eqref{eq:tsn} to stay identity-discriminative even when single frames degrade. We
therefore supervise each frame on its own in addition to the pooled average, with one shared ArcFace
classifier~\cite{deng2019arcface} and identity label $y$, giving the dual-consensus objective:
\begin{equation}
  \mathcal{L}
  \;=\;
  \underbrace{\mathcal{L}_{\mathrm{arc}}\!\big(\phi(\bar{\mathbf{f}}),\,y\big)}_{\mathcal{L}_{\mathrm{cons}}}
  \;+\;\alpha\cdot
  \underbrace{\frac{1}{K}\sum_{t=1}^{K}\mathcal{L}_{\mathrm{arc}}\!\big(\phi(\mathbf{f}^{(t)}),\,y\big)}_{\mathcal{L}_{\mathrm{frame}}}.
  \label{eq:dual}
\end{equation}
Its two terms apply the same ArcFace loss at two levels: $\mathcal{L}_{\mathrm{cons}}$ to the pooled clip
embedding $\phi(\bar{\mathbf{f}})$ that the global path compares, and $\mathcal{L}_{\mathrm{frame}}$ to each
per-frame embedding $\phi(\mathbf{f}^{(t)})$, averaged over the $K$ frames and weighted by $\alpha$. Both act on
the global embedding through the shared head $\phi$. The region descriptors carry no separate loss by design:
which regions to trust is decided at matching time by the transport (\S\ref{subsec:region}), so supervising them
here would duplicate the region path. The objective therefore sharpens the global view that the fusion depends on
(\S\ref{subsec:ablation-exp}).

\subsection{Region Path: Entropic Optimal-Transport Matching}
\label{subsec:region}
\begin{figure*}[t!]\centering
  \includegraphics[width=\textwidth]{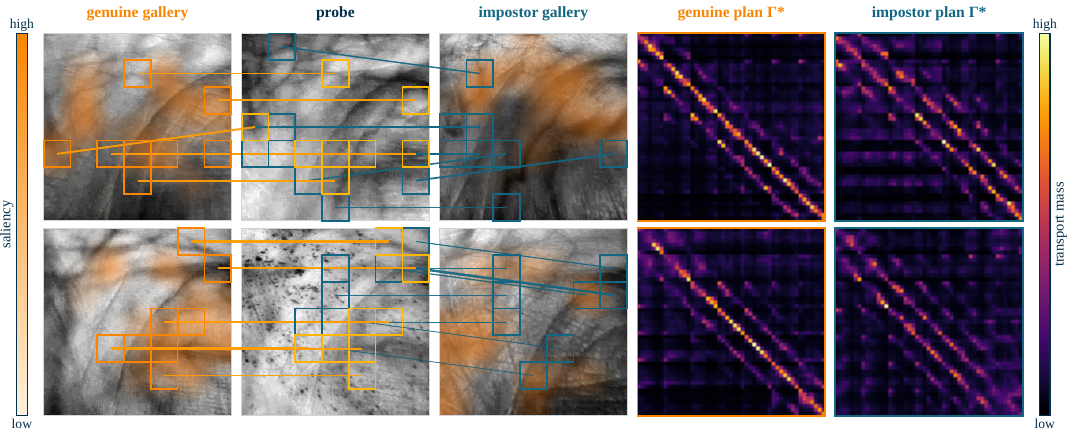}
  \caption{\textbf{Region Matching by Entropic Optimal Transport.} Each row is a real degraded probe (top:
  warm; bottom: dirty), shown between its genuine gallery (left) and the impostor that the global path ranks
  first (right). Boxes and links mark the eight strongest region assignments to each side, and the gallery
  images are shaded by saliency (scale bar at far left); darker shading marks the more discriminative regions.
  The two rightmost panels are the transport plans toward the genuine and the impostor gallery, brighter entries
  carrying more mass. Assignments toward the genuine gallery land coherently on high-saliency regions that
  survive the degradation, and its plan is visibly peaked; assignments toward the impostor are scattered, and
  its plan diffuse.}
  \label{fig:regionot}
\end{figure*}
With the two views in hand, the region path scores each gallery by how coherently its region set aligns with the probe's, region by region.
We cast this as optimal transport (OT). Instead of picking a single best-matching region, OT seeks a soft,
mass-balanced assignment between the two $R$-region grids, so transport mass flows off corrupted regions and onto
the surviving ones. The mass balance also bounds any single region pair to at most $1/R$ of the transported mass,
so a high score requires agreement across many regions. Fig.~\ref{fig:regionot} shows why this rescues degraded probes that the global
path loses. The strongest assignments to the genuine gallery land coherently on high-saliency regions that
survived the degradation. Those to the impostor the global path ranked first are scattered, and the genuine
transport plan is correspondingly peaked. The region score thus places the genuine identity first even when the
pooled global embedding fails.

The transport cost between gallery region $i$ and probe region $j$ is the cosine distance
\begin{equation}
  \mathbf{M}_{ij}(G) \;=\; 1 - \hat{\mathbf{g}}_i(G)^{\!\top}\hat{\mathbf{p}}_j ,
  \qquad i,j\in\{1,\dots,R\},
  \label{eq:cost}
\end{equation}
where $G$ indexes an enrolled gallery identity and $R{=}P^{2}$ is the number of regions per grid
(\S\ref{subsec:multiframe}). The transport plan $\Gamma^{\star}(G)$ is the soft assignment that carries each
probe region's mass onto the gallery's regions at the least total cost. We obtain it from the entropy-regularized
problem with gallery marginal $\mathbf{a}(G)$ and probe marginal $\mathbf{b}$,
\begin{equation}
  \Gamma^{\star}(G)=\operatorname*{arg\,min}_{\Gamma\in U(\mathbf{a}(G),\,\mathbf{b})}\;
  \langle \Gamma,\mathbf{M}(G)\rangle \;-\; \varepsilon\, H(\Gamma),
  \label{eq:ot}
\end{equation}
where the transport polytope
$U(\mathbf{a},\mathbf{b})=\{\Gamma\ge 0:\Gamma\mathbf{1}{=}\mathbf{a},\,\Gamma^{\!\top}\mathbf{1}{=}\mathbf{b}\}$
couples the two marginals, $H(\Gamma)=-\sum_{ij}\Gamma_{ij}(\log\Gamma_{ij}-1)$ is the entropy term, and
$\varepsilon$ is the regularization weight. Problem~\eqref{eq:ot} is solved by $L$ Sinkhorn iterations in the log
domain~\cite{cuturi2013sinkhorn}. The region score is the transported cosine similarity
\begin{equation}
  S_{\mathrm{reg}}(G)\;=\;\big\langle \Gamma^{\star}(G),\,\mathbf{1}-\mathbf{M}(G)\big\rangle.
  \label{eq:sreg}
\end{equation}
where $\langle\cdot,\cdot\rangle$ is the elementwise (Frobenius) inner product, so $S_{\mathrm{reg}}$
accumulates similarity exactly where the plan places mass. The probe marginal is uniform,
$\mathbf{b}=\tfrac1R\mathbf{1}$, and the last Sinkhorn update fixes it exactly. The plan's total mass is thus one,
and $S_{\mathrm{reg}}$ equals one minus the transported cost: the similarity and cost forms coincide.

\para{Gallery Saliency Marginal.}
Not all regions are equally identifying, and the marginal $\mathbf{a}(G)$ controls how much transport budget each
gallery region may claim (the tint on the gallery images in Fig.~\ref{fig:regionot}). Spending that budget on
discriminative regions is more robust than spreading it uniformly. We therefore compute, once at enrollment and
from the clean gallery descriptors alone (unaffected by probe corruption), a per-region discriminativeness score
\begin{equation}
  d_i(G)\;=\;\frac{1}{|\mathcal{G}_{\mathrm{gal}}|-1}
  \sum_{G'\neq G}\Big(1-\hat{\mathbf{g}}_i(G)^{\!\top}\hat{\mathbf{g}}_i(G')\Big),
  \label{eq:disc}
\end{equation}
the mean cosine distance of region $i$ to the same region across all other galleries
$\mathcal{G}_{\mathrm{gal}}$. A region that looks alike across identities is uninformative ($d_i$ small), whereas
one that varies is identifying ($d_i$ large). The gallery marginal up-weights the latter,
\begin{equation}
  a_i(G)\;\propto\;\big(\max(0,d_i(G))+\delta\big)^{\gamma},
  \qquad \textstyle\sum_i a_i(G)=1,
  \label{eq:sal}
\end{equation}
with sharpness $\gamma$ and floor $\delta$. Substituting $\mathbf{a}(G)$ from~\eqref{eq:sal} into~\eqref{eq:ot}
concentrates the transport plan on the high-saliency regions that carry the genuine match in Fig.~\ref{fig:regionot}.

\subsection{Complementary Fusion}
\label{subsec:fusion}
The global path is the trivial holistic scorer that the fusion combines with the region score: a plain cosine
between the aggregated embeddings,
\begin{equation}
  S_{\mathrm{glob}}(G)\;=\;\mathbf{u}_p^{\!\top}\mathbf{u}_G ,
  \label{eq:sglob}
\end{equation}
which stays discriminative on clean captures and on probes whose local patches are individually ambiguous but
whose overall pattern still matches, the cases the region path finds hardest. The two scores are not directly
comparable. $S_{\mathrm{glob}}$ is a raw cosine, whereas $S_{\mathrm{reg}}$ is a transported similarity on a
different scale, and their spreads over the gallery set differ from probe to probe. We therefore standardize each
score per probe over the gallery set, a standard score-normalization step,
\begin{equation}
  z\!\left[S\right]\!(G)\;=\;\frac{S(G)-\mu_S}{\sigma_S},
  \label{eq:z}
\end{equation}
where $\mu_S$ and $\sigma_S$ are the mean and standard deviation of the gallery scores $\{S(G')\}_{G'}$, and we
fuse the two standardized scores by a weighted sum,
\begin{equation}
  S(G)\;=\;z\!\left[S_{\mathrm{reg}}\right]\!(G)\;+\;\lambda\, z\!\left[S_{\mathrm{glob}}\right]\!(G),
  \label{eq:fuse}
\end{equation}
with a single weight $\lambda$ held fixed across all surfaces and databases (\S\ref{subsec:setup-exp}); the
identity is decided by $\operatorname*{arg\,max}_G S(G)$. The per-probe standardization is what renders the two
heterogeneous scores commensurable, and the fusion adds no trained parameters. It reads the probe's scores over
the enrolled gallery, which an enrolment-based deployment already holds at match time. The two paths fail on different
probes: the region path when local patches mislead the transport, the global path when corruption dominates the
embedding. Fusing them therefore beats either alone (\S\ref{subsec:ablation-exp}). The whole matcher adds no learned parameters
and runs entirely at test time.

\section{Experiments}
\label{sec:experiments}

The experiments run under the common protocol of \S\ref{subsec:setup-exp}. \S\ref{subsec:main-exp} benchmarks
the method against twenty-one recognizers on CUP, then dissects the gain: component-by-component ablations,
sensitivity to test-time settings, the frame budget behind the temporal consensus, and a subject-disjoint
benchmark. \S\ref{subsec:general-exp} tests generalization across frozen backbones and four public
databases, and \S\ref{subsec:fairness-exp} closes with a preliminary fairness audit.

\begin{table*}[t!]\centering
\caption{\textbf{Summary of the Five Palm-Vein Evaluation Databases.} Counts are distinct identities (palms); left and right palms are counted as distinct identities. Sample counts are 60-frame video clips for CUP and still images for the public databases.}
\label{tab:datasets}
\footnotesize
\setlength{\tabcolsep}{5pt}
\renewcommand{\arraystretch}{1.3}
\begin{tabular*}{\textwidth}{@{\extracolsep{\fill}}l c c c c c}
\toprule
Database & Identities & \shortstack{Total\\Samples} & \shortstack{Train\\(ID / \#samples)} & \shortstack{Test\\(ID / \#samples)} & \shortstack{Authentication\\Pairs} \\
\midrule
CUP (ours) & 210 & 5{,}049 (videos) & 106 / 2{,}376 & 104 / 2{,}673 & 267{,}176 \\
SCUT~\cite{luo2024palm} & 1{,}100 & 11{,}000 (images) & 550 / 5{,}500 & 550 / 5{,}500 & 2{,}722{,}500 \\
TJ-PV~\cite{zhang2018palmprint} & 600 & 12{,}000 (images) & 300 / 6{,}000 & 300 / 6{,}000 & 1{,}710{,}000 \\
CASIA (850\,nm)~\cite{hao2008multispectral} & 200 & 1{,}200 (images) & 100 / 600 & 100 / 600 & 50{,}000 \\
VERA~\cite{tome2015vulnerability} & 220 & 2{,}200 (images) & 110 / 1{,}100 & 110 / 1{,}100 & 108{,}900 \\
\bottomrule
\end{tabular*}
\end{table*}
\subsection{Experimental Setup}
\label{subsec:setup-exp}
\paragraph{Data.}
We evaluate on five contactless palm-vein databases (Table~\ref{tab:datasets}): our video benchmark CUP and four
public single-image sets, SCUT~\cite{luo2024palm}, TJ-PV~\cite{zhang2018palmprint},
CASIA~\cite{hao2008multispectral}, and VERA~\cite{tome2015vulnerability}. CUP provides 210 palms and 5{,}049
two-second clips under four surface conditions. The single-image sets have no temporal data, so they test only the
spatial part of the matcher, for which one frame suffices. Table~\ref{tab:datasets} lists the split for each. On
CUP we split by hand: the 106 left palms train and the 104 right palms test. The same participants thus appear
on both sides, with every demographic and physiological group, as the fairness analysis of
\S\ref{subsec:fairness-exp} requires. A subject-disjoint benchmark (Table~\ref{tab:leakage}) repeats the
comparison with no shared subjects.

\paragraph{Protocol.}
All results use an open-set protocol: the training and test palms are disjoint, so no palm seen in training
appears at test. The same participants recur across the two splits by design. Every recognizer is trained on the mixed-surface training split and evaluated on each surface
condition separately. At test time we form a gallery--probe setup. Each of the $G$ test subjects is enrolled once,
from a favorable-surface clip on CUP and from the first
image on each public set; every remaining sample serves as a probe. Enrollment therefore uses a favorable
capture while every degraded sample appears as a probe, the setting a supervised enrolment session affords. Within a surface condition, a palm's gallery and probe clips may come from the same recording. We match each probe against all $G$
templates rather than a balanced subset of pairs, which gives
\begin{equation}
N_{\mathrm{pair}}=\underbrace{P}_{\text{genuine}}+\underbrace{P\,(G{-}1)}_{\text{impostor}}
=P\,G,\qquad P=N_{\mathrm{test}}-G,
\label{eq:pairs}
\end{equation}
authentication pairs for $G$ enrolled subjects and $P$ probes among $N_{\mathrm{test}}$ test samples
(Table~\ref{tab:datasets}). Scoring every pair keeps the low-FAR operating points well sampled and avoids any
pairing bias. Because EER, TAR, and Rank-1 are within-class rates, they do not depend on the
genuine-to-impostor ratio. We report three metrics: the equal error rate (EER, $\downarrow$), the operating point
at which the false accept and false reject rates coincide, which summarizes verification error; TAR@FAR${=}0.01$
($\uparrow$), the true accept rate at a false accept rate of $0.01$, which reads verification at a strict
threshold; and Rank-1 accuracy ($\uparrow$), the fraction of probes whose top-ranked gallery is correct, which
measures identification.

\paragraph{Baselines.}
We compare against twenty-one recognizers chosen to cover a wide range of architectures and to include strong
reference points: the vascular-specific recognizers RSNet~\cite{luo2025rsnet}, AMPVNet~\cite{luo2024palm},
MDNet~\cite{kuzu2021loss}, FCPVN~\cite{ma2023focal}, and GSCL-FusionAug~\cite{ou2024gscl}; the generic
convolutional backbones EfficientNet-B0~\cite{tan2019efficientnet} and ResNet-18~\cite{he2016deep}; the
transformers Swin-T~\cite{liu2021swin}, SwinV2-T~\cite{liu2022swinv2}, MaxViT-T~\cite{tu2022maxvit}, and
DeiT3-S~\cite{touvron2022deit3}; the end-to-end video models TSN-R18~\cite{wang2016tsn}, R3D-18~\cite{hara2018can},
R(2+1)D-18~\cite{tran2018r2plus1d}, C3D~\cite{tran2015c3d}, Swin3D-T and Swin3D-S~\cite{liu2022videoswin}, and
MViTv2-S~\cite{li2022mvitv2}, which take the same multi-frame input as our method; and the aggregation methods
from face video recognition, NAN~\cite{yang2017nan}, GhostVLAD~\cite{zhong2018ghostvlad}, and
CAFace~\cite{kim2022caface}.

\paragraph{Implementation.}
The shared training schedule uses AdamW (SGD for MDNet, following its reference) with a per-epoch cosine schedule
and weight decay $10^{-4}$. Learning rates are $3{\times}10^{-4}$ for the convolutional backbones and
$1{\times}10^{-4}$ for the transformers, which are data-hungry and start from pretrained checkpoints. Batch
size is $64$ for the static models and $12$ for the video models. Inputs are $224{\times}224$ single-channel ROIs, replicated to three channels with ImageNet normalization,
a random affine transform (rotation $\pm12^\circ$, scale $[0.9,1.1]$, translation $\pm6\%$, applied with
$p{=}0.5$), and the random gamma augmentation (RGA, $\gamma\in[0.7,1.4]$) introduced in~\cite{luo2024palm}.

Several baselines release no public implementation; we reimplement them from their papers as faithfully as we can. Each recognizer otherwise follows its source paper, including the choice of head and metric; the transformers
are initialized from ImageNet. Before finetuning on CUP, every backbone is pretrained on the public palm-vein
datasets of Table~\ref{tab:datasets}, whose larger set of identities gives stronger vein features. This
pretraining applies to all backbones; for the video models we assemble the same data into pseudo-video sequences. Each
multi-frame model uses $8$ frames per clip, a budget we ablate in \S\ref{subsec:multiframe-exp}. For the
aggregation methods, as for our own model, we use EfficientNet-B0 as a general-purpose backbone.

Our method finetunes the backbone with the dual-consensus objective at learning rate $3{\times}10^{-5}$ and
$\alpha{=}2.0$. Its test-time matcher uses a $7{\times}7$ grid ($R{=}49$), entropic OT with $\varepsilon{=}0.1$
and $L{=}50$ Sinkhorn iterations, a saliency marginal with $\gamma{=}2$ and $\delta{=}0.05$, and a fusion weight
of $\lambda{=}0.5$, held fixed across every surface and database. The grid is set by the backbone's
$7{\times}7$ feature map, and $\varepsilon{=}0.1$ and $L{=}50$ are the usual entropic-transport defaults. We
fixed the remaining values ($\gamma$, $\delta$, $\lambda$, $K$) in advance rather than searching over them. We train the recognizer with five random
seeds and report their mean; every individual seed beats all baselines on all four surfaces. The sensitivity
analyses below fix a single representative seed, on which the full model scores a $3.29\%$ mean EER.

\begin{table*}[t!]\centering
\caption{\textbf{Open-Set Performance on CUP Across Surface Conditions.} Per-surface EER ($\downarrow$),
TAR@FAR${=}0.01$ ($\uparrow$), and Rank-1 ($\uparrow$), all in \%, with each method's inference parameter count and per-probe encoding FLOPs at its frame budget. Bold marks the best per column and underline the best baseline.}
\label{tab:main}
\footnotesize
\setlength{\tabcolsep}{3pt}
\renewcommand{\arraystretch}{1.3}
\begin{tabular*}{\textwidth}{@{\extracolsep{\fill}}c l c c cccc cccc cccc}
\toprule
& \multirow{2}{*}{Method} & \multirow{2}{*}{\shortstack{Params\\(M)}} & \multirow{2}{*}{\shortstack{FLOPs\\(G)}} & \multicolumn{4}{c}{EER \% $\downarrow$} & \multicolumn{4}{c}{TAR@FAR${=}0.01$ $\uparrow$} & \multicolumn{4}{c}{Rank-1 \% $\uparrow$} \\
\cmidrule(lr){5-8}\cmidrule(lr){9-12}\cmidrule(lr){13-16}
& & & & base. & warm & wet & dirty & base. & warm & wet & dirty & base. & warm & wet & dirty \\
\midrule
\multirow{5}{*}{\rotatebox[origin=c]{90}{\scriptsize\shortstack{Palm-vein\\specific}}}
& RSNet~\cite{luo2025rsnet} & 4.0 & 0.21 & 3.33 & 7.73 & 5.67 & 16.50 & 94.43 & 82.80 & 85.94 & 52.58 & 93.59 & 80.89 & 84.30 & 52.23 \\
& AMPVNet~\cite{luo2024palm} & 0.7 & 0.12 & 3.90 & 8.06 & 5.86 & 16.59 & 91.17 & 78.10 & 81.47 & 47.52 & 91.59 & 77.47 & 79.88 & 47.62 \\
& MDNet~\cite{kuzu2021loss} & 27.6 & 7.73 & 4.44 & 7.27 & 6.80 & 17.41 & 89.47 & 79.72 & 79.43 & 41.45 & 87.60 & 75.37 & 76.55 & 38.30 \\
& FCPVN~\cite{ma2023focal} & 11.3 & 1.81 & 8.55 & 13.22 & 12.37 & 22.72 & 74.23 & 51.76 & 51.71 & 20.60 & 70.60 & 52.35 & 50.37 & 20.90 \\
& GSCL-FusionAug~\cite{ou2024gscl} & 11.3 & 1.81 & 5.45 & 8.65 & 6.99 & 20.92 & 88.38 & 70.82 & 76.15 & 30.53 & 85.30 & 68.04 & 73.17 & 27.97 \\
\midrule
\multirow{2}{*}{\rotatebox[origin=c]{90}{\scriptsize CNN}}
& EfficientNet-B0~\cite{tan2019efficientnet} & 4.3 & 0.39 & 4.35 & 9.24 & 7.86 & 20.15 & 91.83 & 76.54 & 78.24 & 39.10 & 90.20 & 72.73 & 75.71 & 34.89 \\
& ResNet-18~\cite{he2016deep} & 11.3 & 1.81 & 5.12 & 9.39 & 7.14 & 22.42 & 90.56 & 75.51 & 80.18 & 39.10 & 88.38 & 72.29 & 76.75 & 37.44 \\
\midrule
\multirow{4}{*}{\rotatebox[origin=c]{90}{\scriptsize\shortstack{Vision\\transformer}}}
& Swin-T~\cite{liu2021swin} & 27.7 & 4.49 & 3.41 & 8.18 & 7.19 & 15.79 & 91.65 & 71.85 & 78.69 & 44.96 & 88.20 & 68.48 & 71.98 & 42.56 \\
& SwinV2-T~\cite{liu2022swinv2} & 27.8 & 4.94 & 4.90 & 10.26 & 9.39 & 17.62 & 88.57 & 66.42 & 70.94 & 35.49 & 88.02 & 65.98 & 67.36 & 36.99 \\
& MaxViT-T~\cite{tu2022maxvit} & 30.3 & 5.56 & 3.27 & 8.66 & 6.11 & 17.26 & 92.92 & 80.06 & 83.01 & 44.51 & 92.74 & 77.27 & 77.50 & 41.50 \\
& DeiT3-S~\cite{touvron2022deit3} & 21.8 & 4.24 & 11.64 & 23.56 & 20.57 & 32.84 & 65.34 & 37.39 & 41.88 & 15.79 & 65.52 & 37.24 & 40.39 & 15.49 \\
\midrule
\multirow{7}{*}{\rotatebox[origin=c]{90}{\scriptsize\shortstack{End-to-end\\video}}}
& TSN-R18~\cite{wang2016tsn} & 11.3 & 14.5 & 1.99 & 4.08 & 3.85 & 17.14 & 96.91 & 90.91 & 91.51 & 51.73 & 94.92 & 87.10 & 90.46 & 48.27 \\
& R3D-18~\cite{hara2018can} & 33.3 & 20.3 & 2.52 & 5.71 & 4.91 & 16.28 & 94.74 & 87.39 & 87.18 & 53.53 & 96.01 & 85.78 & 86.14 & 48.27 \\
& R(2+1)D-18~\cite{tran2018r2plus1d} & 31.5 & 20.3 & \underline{1.11} & 4.27 & 3.00 & 14.27 & \underline{98.73} & 88.86 & 93.00 & 49.02 & 96.55 & 84.02 & 90.76 & 45.26 \\
& C3D~\cite{tran2015c3d} & 27.8 & 32.9 & 2.94 & 7.64 & 5.83 & 14.75 & 94.56 & 81.96 & 85.10 & 49.47 & 93.65 & 79.62 & 80.77 & 48.27 \\
& Swin3D-T~\cite{liu2022videoswin} & 28.1 & 19.7 & 5.07 & 10.40 & 7.56 & 14.62 & 81.85 & 64.37 & 63.79 & 50.08 & 75.86 & 64.37 & 63.64 & 47.37 \\
& Swin3D-S~\cite{liu2022videoswin} & 49.7 & 37.8 & 5.26 & 8.50 & 7.63 & 14.60 & 87.66 & 75.22 & 70.94 & 51.58 & 83.12 & 70.09 & 66.92 & 51.28 \\
& MViTv2-S~\cite{li2022mvitv2} & 34.5 & 64.2 & 4.36 & 8.34 & 7.48 & 13.08 & 88.38 & 73.75 & 71.68 & 51.73 & 82.40 & 66.28 & 64.08 & 48.87 \\
\midrule
\multirow{3}{*}{\rotatebox[origin=c]{90}{\scriptsize\shortstack{Face video\\recognition}}}
& NAN~\cite{yang2017nan} & 6.0 & 3.08 & 1.63 & \underline{3.37} & 2.53 & \underline{11.93} & 97.46 & 92.96 & 95.08 & \underline{61.05} & \underline{98.00} & \underline{90.76} & 93.14 & \underline{59.10} \\
& CAFace~\cite{kim2022caface} & 6.0 & 3.08 & 1.47 & \underline{3.37} & \underline{2.09} & 12.79 & 98.19 & \underline{93.11} & \underline{95.53} & 60.45 & 97.82 & 88.86 & \underline{94.19} & 57.74 \\
& GhostVLAD~\cite{zhong2018ghostvlad} & 17.5 & 3.09 & 1.63 & 4.86 & 3.58 & 14.61 & 98.00 & 89.59 & 90.91 & 53.08 & 96.91 & 85.04 & 86.74 & 50.83 \\
\midrule
& Ours & 4.3 & 3.08 & \textbf{0.26} & \textbf{2.14} & \textbf{1.85} & \textbf{9.56} & \textbf{99.96} & \textbf{97.12} & \textbf{97.26} & \textbf{71.94} & \textbf{99.35} & \textbf{94.43} & \textbf{95.29} & \textbf{65.05} \\
\bottomrule
\end{tabular*}\end{table*}
\subsection{Results and Analysis on CUP}
\label{subsec:main-exp}

Table~\ref{tab:main} reports the full benchmark: our method attains the lowest EER, the highest
TAR@FAR${=}0.01$, and the highest Rank-1 on every surface. The gap to the field is widest exactly where a
single global descriptor breaks down: even the best baseline on the dirty condition (NAN, $11.93\%$ EER)
stays far above our $9.56\%$. Two patterns organize the baselines. First, the multi-frame recognizers are the most robust
of the prior methods on the degraded surfaces, with the frame-aggregation models (NAN, CAFace) leading the
end-to-end video models. Second, the advantage over them comes from the matcher rather than the backbone: NAN
and CAFace share our EfficientNet-B0, yet our region-OT matcher surpasses them on every surface and metric.

\label{subsec:ablation-exp}
\begin{table*}[t!]\centering
\caption{\textbf{Step-by-Step Ablation of Our Method.}}
\label{tab:ablation-steps}
\footnotesize
\setlength{\tabcolsep}{4pt}
\renewcommand{\arraystretch}{1.3}
\begin{tabular*}{\textwidth}{@{\extracolsep{\fill}}l cccc cccc cccc}
\toprule
\multirow{2}{*}{Stage} & \multicolumn{4}{c}{EER \% $\downarrow$} & \multicolumn{4}{c}{TAR@FAR${=}0.01$ $\uparrow$} & \multicolumn{4}{c}{Rank-1 \% $\uparrow$} \\
\cmidrule(lr){2-5}\cmidrule(lr){6-9}\cmidrule(lr){10-13}
 & base. & warm & wet & dirty & base. & warm & wet & dirty & base. & warm & wet & dirty \\
\midrule
backbone ($K{=}8$ frames) & 2.00 & 4.25 & 3.41 & 16.80 & 97.46 & 89.15 & 90.91 & 51.58 & 96.37 & 89.15 & 90.16 & 48.12 \\
+ fusion & 0.55 & 3.38 & 2.07 & 12.14 & 99.64 & 95.02 & 96.87 & 63.01 & 98.00 & 90.47 & 93.59 & 58.05 \\
+ saliency & 0.52 & 3.04 & \textbf{1.78} & 12.03 & 99.82 & 95.45 & \textbf{97.47} & 64.66 & 98.19 & 91.20 & 94.19 & 60.45 \\
+ dual-consensus & \textbf{0.26} & \textbf{2.14} & 1.85 & \textbf{9.56} & \textbf{99.96} & \textbf{97.12} & 97.26 & \textbf{71.94} & \textbf{99.35} & \textbf{94.43} & \textbf{95.29} & \textbf{65.05} \\
\bottomrule
\end{tabular*}
\end{table*}
\begin{figure}[t!]\centering
  \includegraphics[width=\columnwidth]{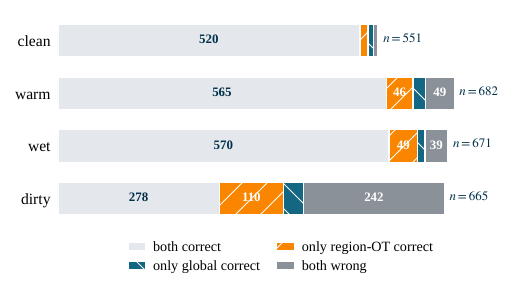}
  \caption{\textbf{Complementarity of the Two Views per Surface.} Share of CUP probes on which each view ranks the
  genuine identity first, split into both correct, only region-OT correct, only global correct, and both wrong.}
  \label{fig:complementarity}
\end{figure}

\begin{table*}[t!]\centering
\caption{\textbf{Test-Time Transfer to Frozen Backbones.} Each cell reads left to right: the backbone's original single-frame performance, with the $K{=}8$ temporal consensus added under the same cosine scoring, and with the full matcher on top. Bold marks the best value.}
\label{tab:transfer}
\footnotesize
\setlength{\tabcolsep}{2pt}
\renewcommand{\arraystretch}{1.35}
\begin{tabular*}{\textwidth}{@{\extracolsep{\fill}}cccccc}
\toprule
Metric & Backbone & baseline & warm & wet & dirty \\
\midrule
\multirow{4}{*}{EER \% $\downarrow$} & RSNet~\cite{luo2025rsnet} & 3.33\,$\to$\,2.54\,$\to$\,\textbf{1.27} & 7.73\,$\to$\,6.62\,$\to$\,\textbf{4.69} & 5.67\,$\to$\,4.77\,$\to$\,\textbf{3.28} & 16.50\,$\to$\,14.58\,$\to$\,\textbf{11.62} \\
 & AMPVNet~\cite{luo2024palm} & 3.90\,$\to$\,3.45\,$\to$\,\textbf{2.36} & 8.06\,$\to$\,5.90\,$\to$\,\textbf{5.58} & 5.86\,$\to$\,4.62\,$\to$\,\textbf{4.20} & 16.59\,$\to$\,15.62\,$\to$\,\textbf{12.03} \\
 & MaxViT-T~\cite{tu2022maxvit} & 3.27\,$\to$\,2.54\,$\to$\,\textbf{1.45} & 8.66\,$\to$\,7.93\,$\to$\,\textbf{5.27} & 6.11\,$\to$\,5.22\,$\to$\,\textbf{3.12} & 17.26\,$\to$\,16.82\,$\to$\,\textbf{13.07} \\
 & Swin-T~\cite{liu2021swin} & 3.41\,$\to$\,2.90\,$\to$\,\textbf{1.45} & 8.18\,$\to$\,7.92\,$\to$\,\textbf{6.31} & 7.19\,$\to$\,6.91\,$\to$\,\textbf{4.62} & 15.79\,$\to$\,15.21\,$\to$\,\textbf{12.04} \\
\midrule
\multirow{4}{*}{\shortstack{TAR@FAR\\${=}0.01$ $\uparrow$}} & RSNet~\cite{luo2025rsnet} & 94.43\,$\to$\,96.01\,$\to$\,\textbf{98.55} & 82.80\,$\to$\,85.04\,$\to$\,\textbf{91.94} & 85.94\,$\to$\,88.38\,$\to$\,\textbf{93.29} & 52.58\,$\to$\,55.34\,$\to$\,\textbf{67.67} \\
 & AMPVNet~\cite{luo2024palm} & 91.17\,$\to$\,94.37\,$\to$\,\textbf{96.91} & 78.10\,$\to$\,81.38\,$\to$\,\textbf{90.18} & 81.47\,$\to$\,84.05\,$\to$\,\textbf{89.72} & 47.52\,$\to$\,47.97\,$\to$\,\textbf{69.47} \\
 & MaxViT-T~\cite{tu2022maxvit} & 92.92\,$\to$\,96.73\,$\to$\,\textbf{98.37} & 80.06\,$\to$\,83.14\,$\to$\,\textbf{89.30} & 83.01\,$\to$\,85.84\,$\to$\,\textbf{93.89} & 44.51\,$\to$\,46.62\,$\to$\,\textbf{57.74} \\
 & Swin-T~\cite{liu2021swin} & 91.65\,$\to$\,93.10\,$\to$\,\textbf{98.00} & 71.85\,$\to$\,74.63\,$\to$\,\textbf{85.78} & 78.69\,$\to$\,77.65\,$\to$\,\textbf{88.38} & 44.96\,$\to$\,47.67\,$\to$\,\textbf{61.65} \\
\midrule
\multirow{4}{*}{Rank-1 \% $\uparrow$} & RSNet~\cite{luo2025rsnet} & 93.59\,$\to$\,94.19\,$\to$\,\textbf{94.92} & 80.89\,$\to$\,85.48\,$\to$\,\textbf{87.83} & 84.30\,$\to$\,86.44\,$\to$\,\textbf{87.33} & 52.23\,$\to$\,52.63\,$\to$\,\textbf{60.15} \\
 & AMPVNet~\cite{luo2024palm} & 91.59\,$\to$\,93.65\,$\to$\,\textbf{94.19} & 77.47\,$\to$\,82.11\,$\to$\,\textbf{83.72} & 79.88\,$\to$\,\textbf{83.46}\,$\to$\,83.01 & 47.62\,$\to$\,49.32\,$\to$\,\textbf{63.01} \\
 & MaxViT-T~\cite{tu2022maxvit} & 92.74\,$\to$\,95.64\,$\to$\,\textbf{96.91} & 77.27\,$\to$\,80.94\,$\to$\,\textbf{83.58} & 77.50\,$\to$\,80.92\,$\to$\,\textbf{86.59} & 41.50\,$\to$\,43.91\,$\to$\,\textbf{51.73} \\
 & Swin-T~\cite{liu2021swin} & 88.20\,$\to$\,92.01\,$\to$\,\textbf{94.19} & 68.48\,$\to$\,73.61\,$\to$\,\textbf{80.35} & 71.98\,$\to$\,76.01\,$\to$\,\textbf{82.41} & 42.56\,$\to$\,47.22\,$\to$\,\textbf{53.68} \\
\bottomrule
\end{tabular*}
\end{table*}

\paragraph{Ablations.}
We now dissect where this gain comes from by building the method one component at a time on a frozen
EfficientNet-B0 (Table~\ref{tab:ablation-steps}). The starting row is the frozen backbone with its $K{=}8$ frame
average scored by global cosine; the frame budget itself is studied separately below. Fusion and the saliency marginal improve all three metrics on
all four surfaces; dual-consensus training improves the baseline, warm, and dirty conditions. The single largest
step is the fusion of the two views, which are individually insufficient and fail on different probes, so we look
at that complementarity next.

The reduction is steepest on the hardest surfaces. Fusing the global and regional views, with transport under a uniform marginal, cuts the mean EER from
$6.62\%$ to $4.53\%$, a $32\%$ relative drop, and the dirty-condition EER from $16.80\%$ to $12.14\%$. Replacing that uniform marginal with the
gallery saliency prior refines the mean to $4.34\%$. Dual-consensus training gives our final $3.45\%$ across five seeds, again with
the largest absolute gain on the dirty surface ($12.03\%$ to $9.56\%$). We next check how many probes each view gets right on its own (Fig.~\ref{fig:complementarity}). For every test
probe we rank all $104$ gallery templates separately under each view and record which places the genuine template
first. Every probe counts once, so the bars are probes, not pairs. Region-OT alone ranks the genuine palm first on $219$ probes and the global view alone on
$79$. The gap widens with degradation: on the dirty surface region-OT alone rescues $110$ probes against $35$
for the global view, while both miss a hard core of $36\%$. These are ranking outcomes, so they mark where the two
views carry separable information rather than how much of it the fused score keeps, which is the EER reduction
above.

\paragraph{Sensitivity.}
Sweeping each parameter around its default while holding the others
fixed keeps the mean EER within a $3.27\%$--$3.53\%$ band. That band is no wider than the model's own
seed-to-seed variation ($3.45\% \pm 0.16$ over five seeds), so no test-time setting moves the result further
than retraining does. The Sinkhorn
parameters barely register: the mean EER is flat at $3.29\%$ across $L\in\{10,25,50,100\}$ iterations and stays
within $3.29\%$--$3.35\%$ for $\varepsilon\le0.2$, rising only to $3.48\%$ at $\varepsilon{=}0.5$. The saliency
shape is nearly as forgiving: the floor $\delta$ holds within $3.29\%$--$3.31\%$, and the sharpness $\gamma$ moves
the mean EER from $3.27\%$ at $\gamma{=}1$ to $3.53\%$ at $\gamma{=}4$, so our default sits just above the
sweep's best point. We hold $\varepsilon{=}0.1$, $L{=}50$,
$\gamma{=}2$, $\delta{=}0.05$, the saliency marginal, and a $7{\times}7$ grid across every surface and database.

\label{subsec:multiframe-exp}

\begin{figure*}[t!]\centering
  \includegraphics[width=\textwidth]{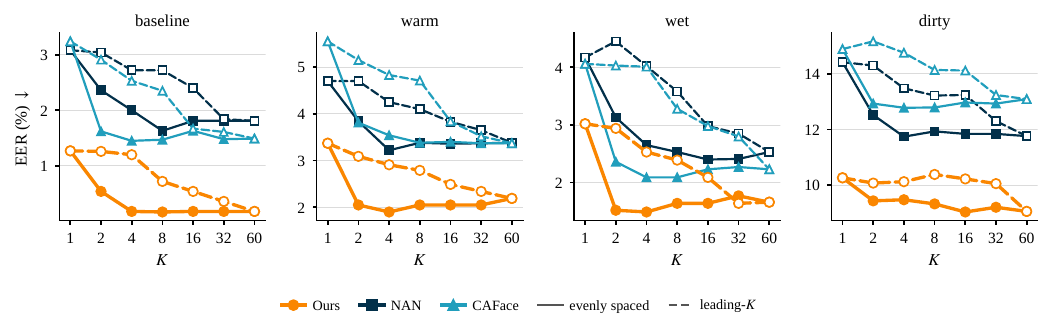}
  \caption{\textbf{Frame Budget and Sampling per Surface.} EER versus the number of probe frames $K$ for our model
  and two multi-frame backbones (NAN~\cite{yang2017nan} and CAFace~\cite{kim2022caface}), under evenly spaced (solid, filled) and leading-$K$ (dashed, hollow) sampling, shown
  for each surface condition. Evenly spaced sampling is superior throughout.}
  \label{fig:frames}
\end{figure*}

\paragraph{Frame budget.}
We test two sampling strategies, evenly spaced and leading-$K$, under varying frame budgets
(Fig.~\ref{fig:frames}). A short clip is enough, provided the frames spread across it. Raising the probe budget
from one frame to a few evenly spaced ones gives almost all of the benefit: the mean EER drops from $4.48\%$ at
$K{=}1$ to $3.26\%$ at $K{=}4$. It then holds in a $3.2\%$--$3.3\%$ band out to $K{=}60$; our $K{=}8$ setting
sits at $3.29\%$. Coverage matters more than raw count: even spacing beats taking the leading $K$ frames at every
budget ($3.29\%$ against $4.07\%$ at $K{=}8$), consecutive frames being largely redundant. The two schemes coincide
only at the endpoints $K{=}1$ and $K{=}60$, where both span the clip. Beyond a few frames the per-surface curves
wobble by a few tenths of a percent, but this is run-to-run variation rather than structure in $K$. EER is a
single-threshold statistic on a finite, corrupted probe set, so a handful of borderline pairs move it; the
fluctuations are also uncorrelated across the three models. In practice the sensor already returns about sixty frames
over a two-second capture at $30$\,fps, and only a handful spread across the clip are needed. The method thus adds
no acquisition time, and its added cost at deployment is confined to the matching stage.

\paragraph{Deployment cost.}
We quantify that cost per probe on one H100 GPU in fp32. Encoding costs $4.4$\,ms (EfficientNet-B0 over the
$K{=}8$ frames, $3.08$ GFLOPs); the matcher adds $6.6$\,ms, against $0.06$\,ms for a raw cosine.
The matcher is thus the larger component, and its cost has two distinct readings. In FLOPs the matching stage is
$\mathcal{O}(G)$: about $6.5$\,MFLOPs per gallery template, which overtakes the encoding near $G{=}550$ enrolled
identities. In wall clock the growth is sub-linear. The per-template matches are batched on the GPU, so latency
is bounded by the $L{=}50$ sequential Sinkhorn iterations rather than by the gallery scan. It stays near
$6.6$\,ms per probe up to $G{=}550$ and reaches $13.5$\,ms at $G{=}2750$. Total inference therefore stays
around $11$\,ms per probe at realistic enrollment sizes.

\paragraph{Leakage control.}
Because a participant's two hands fall in opposite splits, the training and test palms are disjoint while the same
subjects recur. The two hands share subject-level attributes (skin thickness, body fat, perfusion) that
shape the near-infrared signal, so this split could in principle leak subject-level distribution into the test set.
We therefore repeat the comparison under a subject-disjoint protocol: five random partitions of $52$ training and
$52$ test subjects with no identity overlap, both palms of a subject falling on the same side of the split.
Table~\ref{tab:leakage} reruns the two strongest baselines and the strongest palm-vein-specific one under that
protocol, with the same training recipe as the main table.

Halving the training pool raises absolute error for every method, which reflects data quantity rather than
leakage. The ordering is unchanged and our margin widens: our mean EER is $29\%$ below NAN on the standard split
and $50\%$ below it here, and we lead on every surface. The methods that learn how to aggregate lose the most,
NAN rising by $75\%$ and CAFace by $89\%$ against $24\%$ for our method, which is what a matching stage carrying
no learned parameters would predict.

\begin{table}[t]\centering
\caption{\textbf{Subject-Disjoint Benchmark.} Per-surface EER (\%) averaged over five random $52$/$52$ splits
with zero identity overlap; the mean column carries the spread across splits. Bold marks the best per column.}
\label{tab:leakage}
\footnotesize
\setlength{\tabcolsep}{4pt}
\renewcommand{\arraystretch}{1.2}
\begin{tabular}{@{}lccccc@{}}
\toprule
Method & base. & warm & wet & dirty & mean \\
\midrule
RSNet~\cite{luo2025rsnet} & 4.68 & 9.55 & 8.10 & 22.27 & 11.15\,$\pm$\,0.74 \\
CAFace~\cite{kim2022caface} & 4.70 & 6.38 & 6.93 & 19.18 & 9.30\,$\pm$\,1.27 \\
NAN~\cite{yang2017nan} & 3.18 & 5.66 & 5.66 & 19.64 & 8.54\,$\pm$\,0.70 \\
\midrule
Ours & \textbf{0.18} & \textbf{2.39} & \textbf{1.94} & \textbf{12.65} & \textbf{4.29\,$\pm$\,0.54} \\
\bottomrule
\end{tabular}
\end{table}

\subsection{Generalization}
\label{subsec:general-exp}

\paragraph{Across backbones.}
To test the generalization of the method, we then apply the test-time components to four frozen state-of-the-art
backbones. As shown in Table~\ref{tab:transfer}, each cell reads left to right: the backbone's single-frame
performance from Table~\ref{tab:main}, the same features under the $K{=}8$ temporal consensus, and the full
matcher on top. The improvement is universal: every backbone improves on every surface and every metric, and
the mean EER falls by $29\%$ to $37\%$ without any retraining. The middle column separates the two
components: the temporal consensus alone lowers the mean EER by $5$--$14\%$, and the regional optimal transport
cuts it by a further $18$--$30\%$ on every backbone. The consistency points at the mechanism: pooling
lets a locally corrupted region contaminate the whole descriptor, and region-level matching contains that damage
regardless of the backbone.

\begin{table*}[t!]\centering
\caption{\textbf{Cross-Database Generalization.} Per-database recognition performance before and after the test-time matcher (each cell: global cosine $\to$ full
matcher) for four frozen backbones on the public single-image databases. These databases carry no temporal data,
so the matcher uses only its spatial component. Bold marks the improved side.}
\label{tab:pubdb}
\footnotesize
\setlength{\tabcolsep}{4pt}
\renewcommand{\arraystretch}{1.35}
\begin{tabular*}{\textwidth}{@{\extracolsep{\fill}}cccccc}
\toprule
Metric & Database & RSNet~\cite{luo2025rsnet} & AMPVNet~\cite{luo2024palm} & MaxViT-T~\cite{tu2022maxvit} & Swin-T~\cite{liu2021swin} \\
\midrule
\multirow{4}{*}{EER \% $\downarrow$}
 & CASIA~\cite{hao2008multispectral} & 2.80\,$\to$\,\textbf{1.44} & 2.00\,$\to$\,\textbf{1.00} & 4.40\,$\to$\,\textbf{2.40} & 6.99\,$\to$\,\textbf{4.00} \\
 & SCUT~\cite{luo2024palm} & 1.88\,$\to$\,2.28 & 2.48\,$\to$\,\textbf{2.29} & 1.88\,$\to$\,\textbf{1.13} & 2.89\,$\to$\,\textbf{2.18} \\
 & TJ-PV~\cite{zhang2018palmprint} & 1.32\,$\to$\,\textbf{0.94} & 1.68\,$\to$\,\textbf{1.07} & 1.85\,$\to$\,\textbf{0.84} & 3.27\,$\to$\,\textbf{1.88} \\
 & VERA~\cite{tome2015vulnerability} & 3.24\,$\to$\,\textbf{2.73} & 2.83\,$\to$\,\textbf{2.63} & 3.46\,$\to$\,\textbf{1.82} & 5.05\,$\to$\,\textbf{3.43} \\
\midrule
\multirow{4}{*}{\shortstack{TAR@FAR\\${=}0.01$ $\uparrow$}}
 & CASIA~\cite{hao2008multispectral} & 95.20\,$\to$\,\textbf{97.80} & 97.60\,$\to$\,\textbf{99.00} & 92.00\,$\to$\,\textbf{96.20} & 85.40\,$\to$\,\textbf{93.20} \\
 & SCUT~\cite{luo2024palm} & 97.31\,$\to$\,96.57 & 95.86\,$\to$\,\textbf{96.44} & 97.35\,$\to$\,\textbf{98.83} & 94.49\,$\to$\,\textbf{96.30} \\
 & TJ-PV~\cite{zhang2018palmprint} & 98.37\,$\to$\,\textbf{99.09} & 97.81\,$\to$\,\textbf{98.90} & 97.19\,$\to$\,\textbf{99.21} & 93.14\,$\to$\,\textbf{97.07} \\
 & VERA~\cite{tome2015vulnerability} & 95.05\,$\to$\,\textbf{96.77} & 95.66\,$\to$\,\textbf{96.57} & 93.94\,$\to$\,\textbf{97.98} & 87.58\,$\to$\,\textbf{94.44} \\
\midrule
\multirow{4}{*}{Rank-1 \% $\uparrow$}
 & CASIA~\cite{hao2008multispectral} & 94.40\,$\to$\,\textbf{95.60} & 96.80\,$\to$\,\textbf{97.60} & 91.80\,$\to$\,\textbf{94.80} & 84.40\,$\to$\,\textbf{89.60} \\
 & SCUT~\cite{luo2024palm} & 93.68\,$\to$\,90.10 & 90.99\,$\to$\,90.04 & 93.82\,$\to$\,\textbf{96.10} & 86.14\,$\to$\,\textbf{89.15} \\
 & TJ-PV~\cite{zhang2018palmprint} & 96.58\,$\to$\,\textbf{97.21} & 95.74\,$\to$\,\textbf{97.39} & 94.05\,$\to$\,\textbf{96.98} & 87.32\,$\to$\,\textbf{91.95} \\
 & VERA~\cite{tome2015vulnerability} & 93.43\,$\to$\,93.43 & 95.56\,$\to$\,94.14 & 93.74\,$\to$\,\textbf{96.87} & 85.56\,$\to$\,\textbf{90.00} \\
\bottomrule
\end{tabular*}
\end{table*}

\paragraph{Across databases.}
We also test generalization across data. The four public single-image databases carry no temporal data, so only
the regional optimal transport applies. As shown in Table~\ref{tab:pubdb}, each cell compares global cosine
before and after the matcher on the same frozen backbone, with no per-database finetuning. The matcher lowers the
EER on fifteen of the sixteen database and backbone combinations and raises TAR@FAR${=}0.01$ on the same fifteen;
the sole exception on both is RSNet on SCUT. It cuts each backbone's mean cross-database EER by $20\%$ to $47\%$.
Rank-1 improves on twelve of the sixteen and degrades on three: it moves only when the top match flips, while the
threshold metrics respond to the whole score distribution. These databases have no surface degradation; the gain
here means regional matching also absorbs the natural variation of pose and illumination. The matcher is thus a
general drop-in component, not an artifact of the CUP protocol.

\subsection{User-Group Fairness}
\label{subsec:fairness-exp}
\begin{figure}[t]
  \centering
  \includegraphics[width=\columnwidth]{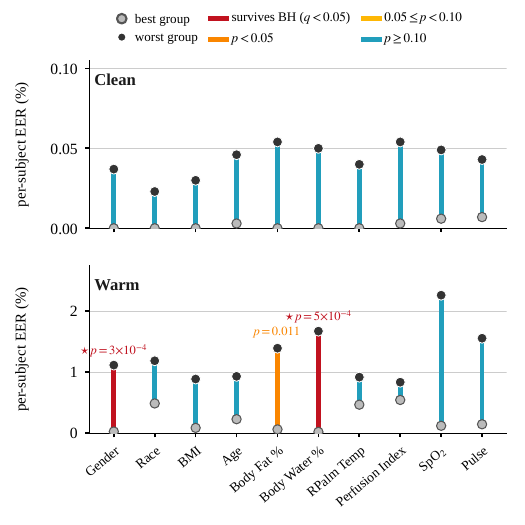}
  \caption{\textbf{Preliminary User-Group Fairness Audit.} Worst-group (dark) versus best-group (gray)
  per-subject EER for ten traits on the clean (top; magnified scale) and warm (bottom) surfaces, with stem
  color marking the rank-permutation $p$ of each gap.}
  \label{fig:fairness}
\end{figure}

We include a preliminary fairness audit rather than a definitive one, because our subject pool is small for
group-wise error analysis. For each subject we compute a per-subject EER from its own genuine and impostor
scores, and partition the subjects by each of ten demographic and palm-physiology traits: gender and race
(Asian vs.\ White, the only categories reaching $\ge3$ subjects in our Asian-majority cohort) use the recorded
labels; BMI uses the standard WHO cut-offs; and the continuous physiology and body-composition traits (age,
body fat, body water, palm temperature, perfusion index, SpO$_2$, pulse) are split into low/mid/high tertiles
at the 33rd and 67th percentiles. A group is retained only if it contains $\ge3$ subjects and a trait only if
it yields $\ge2$ such groups. We restrict the audit to the clean and warm captures. The warm condition is applied
against a measured threshold, a palm temperature above $38.5^\circ$C. On wet and
dirty palms the amount of water or soil that reaches the skin is never measured. How much of it stays there
depends on skin texture and hand geometry, the very traits under audit. A group difference on those surfaces
would mix the contamination with the trait, so we do not read them as evidence about user groups.

Per-subject EER is heavily tied and zero-inflated, since many subjects reach 0\% on the easier captures, which
leaves a difference-of-means statistic under-powered against genuine distributional shifts yet inflated by a
few high-error outliers. We therefore test each group difference with an exact permutation test on the rank
statistic (Mann--Whitney $U$ for two groups, Kruskal--Wallis $H$ for three), drawing the $p$-value from
20{,}000 label permutations so that ties and small samples are handled exactly. Each gap also carries a
$95\%$ percentile bootstrap interval from $5{,}000$ resamples over subjects. Across the twenty
trait--surface tests, two warm-condition traits survive Benjamini--Hochberg correction
(Fig.~\ref{fig:fairness}): body water (worst-to-best gap $1.65$ points, $95\%$ CI $[0.28, 3.41]$,
$p{=}5{\times}10^{-4}$, $q{=}0.005$) and gender (gap $1.09$, CI $[0.25, 2.19]$, $p{=}3{\times}10^{-4}$,
$q{=}0.005$). Both remain significant under the more conservative Bonferroni control. A third, body fat (gap
$1.33$, CI $[0.10, 3.14]$), is significant only before correction ($p{=}0.011$), and no clean-surface trait
approaches significance. Race shows no significant difference on either surface ($p{=}0.56$ clean, $p{=}0.99$
warm). Its warm gap of $0.70$ points carries a wide interval of $[-0.74, 2.41]$, so a difference of that size
cannot be excluded at our sample size. Blood-oxygen saturation shows the largest raw gap ($2.14$ points, CI
$[0.03, 5.87]$) yet is not significant under the rank test ($p{=}0.33$). The gap comes from a few outliers
rather than a shift of the group distribution. A mean-based permutation test on the same data flags it at
$p<0.05$, the spurious call the rank statistic avoids. That the surviving
disparities appear only under the warm condition is expected: on clean palms every group already operates near
the error floor, leaving almost no room for a difference to register, whereas warming lifts error across the
board and opens a wider band of variation in which a group difference can surface.

The two implicated traits are physiologically coherent. Tissue water content influences near-infrared absorption
and scattering, and with them the vein contrast a recognizer sees, so groups differing in body water may image
with slightly different clarity once the surface is stressed; the gender effect is consistent with correlated
body-composition differences and need not reflect an independent factor. We nonetheless treat these as robust
preliminary signals rather than fully characterized biases: they survive multiple-comparison control but rest on
the warm surface alone and on a small subject pool, so we report their direction and corrected significance
while cautioning against over-reading the effect magnitudes. Demographic and body-composition neutrality in
palm-vein recognition should not be assumed, and body water and related composition traits are natural targets
for a larger, adequately powered study.

\section{Conclusion}
\label{sec:conclusion}
We introduced CUP, to our knowledge the first public video-based palm-vein dataset that records each palm under everyday surface degradation with paired physiological and demographic metadata. With it we showed that recognizers which verify reliably on clean palms lose most of their accuracy as the surface degrades. To recover it, we proposed a test-time matcher that adds no learned parameters and pairs the global embedding with a saliency-steered region-level optimal transport and a temporal consensus over the clip. It sets the state of the art on every surface condition of CUP and transfers unchanged to other frozen backbones and to four public single-image datasets.

CUP already spans intra-subject variation far beyond prior public palm-vein datasets, a substantial step toward realistic evaluation. Covering the full deployment distribution remains beyond our protocol, which does not yet cover cross-session use, other forms of contamination, sensor noise, or enrolment from a degraded capture. Future work includes modeling the temporal structure within a clip, extending the protocol across sensors and populations, and treating robustness and fairness jointly. The same temporal signal also carries liveness cues that a single image cannot provide, which we leave to future work. We release CUP and the matcher to support this.

\section*{Acknowledgment}
This research has been facilitated through the generous support of the Mastercard Center for Inclusive Growth.

\bibliographystyle{IEEEtran}
\clearpage
\bibliography{ref}

\vfill

\end{document}